\documentclass[runningheads]{llncs}
\usepackage[T1]{fontenc}
\usepackage{graphicx}
\usepackage{subcaption}
\usepackage{booktabs}
\usepackage{pifont}
\newcommand{\cmark}{\ding{51}}
\newcommand{\xmark}{\ding{55}}

\begin{document}
\title{H-VAEP and H-xT:\\ Valuing Offensive On-the-Ball Actions\\ in Handball by Estimating Probabilities}
\author{Julius Broermann\inst{1} \and
Oliver Mueller\inst{1}, \\
Michael Doering\inst{1,2} \and
Jochen Baumeister\inst{1}
}
\authorrunning{Broermann et al.}
\institute{Paderborn University, Paderborn, Germany \\
\email{\{firstname.lastname\}@uni-paderborn.de}
\vspace{0.5em}\and
SG Flensburg-Handewitt, Flensburg, Germany
}

\titlerunning{H-VAEP and H-xT for Action Valuation in Handball}
\maketitle
\begin{abstract}
Traditional player evaluation in professional handball relies on basic box-score metrics or heuristic indices, which fail to credit the multi-player build-up chain. While football (soccer) analytics has adopted Expected Threat (xT) and Valuing Actions by Estimating Probabilities (VAEP), these event-based action valuation frameworks have not yet been adapted to handball. In this paper, we present the first comprehensive adaptation and evaluation of xT and VAEP for handball, utilizing five seasons of tracking-derived event data from the Handball Bundesliga. We develop Handball-xT (H-xT) using a handball-native court zoning layout, demonstrating via simulations that it is systematically more robust than standard rectangular grids. We optimize Handball-VAEP (H-VAEP) by tailoring its feature space and selecting the context length to limit team-identity leakage. Our evaluation shows that H-VAEP yields exceptionally stable, discriminative, and intuitive player ratings that highlight build-up play. Finally, we release our complete code repository to help professional clubs deploy these models.

\end{abstract}

\section{Introduction}
\label{sec:introduction}
Optimizing player recruitment and game tactics requires a detailed understanding of how individual player actions contribute to winning matches. However, traditional player evaluation in handball remains heavily reliant on basic box-score metrics like goals and assists, or on heuristic, fixed-weight combinations of them such as the Handball Performance Index (HPI)~\cite{hbl_hpi}. While football analytics has successfully transitioned to valuing every on-ball action using Expected Threat (xT)~\cite{singh_introducing_2018} and VAEP~\cite{decroos_actions_2019,decroos_vaep_2021}, professional handball has lacked a comparable framework. Consequently, players who drive the build-up and create scoring opportunities well before the final shot are systematically undervalued. The transfer of football-style action valuation frameworks to handball has historically been hindered by the limitations of scout-collected event data. Traditional box-score statistics record only shots, goals, and assists, omitting the passes and dribbles that constitute the bulk of play. However, the deployment of Local Positioning Systems (LPS) across the Handball Bundesliga (HBL) has resolved this constraint and enables the automatic recognition of event sequences from tracking data, which are made available to the clubs. Yet, a case study of 13 HBL clubs indicates that converting this data into actionable insights remains a major challenge due to scarce analytical resources and specialized expertise~\cite{konzag_sports_2024}.

We address this barrier by developing and evaluating Handball-xT (H-xT) and Handball-VAEP (H-VAEP) using five seasons of HBL tracking-derived event data (2021/22 to 2025/26). To facilitate practical adoption, we release our code repository, including API connectors to data providers.\footnote{The code is available at \url{https://github.com/JuliusBroermann/handballaction}.} Specifically, our contributions are threefold. First, we adapt Expected Threat to handball (H-xT) using a court zoning layout that respects the sport's geometry, validating its robustness over standard rectangular grids via simulation~\cite{van_arem_trade-off_2025}. Second, we adapt VAEP to handball (H-VAEP) by optimizing its features, algorithm, and context length to suit the sport's rapid, high-scoring dynamics while limiting team-identity leakage. Third, we evaluate the resulting ratings under a comprehensive validation framework~\cite{davis_methodology_2024,franks_meta-analytics_2016,sportstutorial2024}, demonstrating their strong face validity, reliability, discrimination, and stability compared to traditional performance indicators.

\section{Related Work}
\label{sec:related_work}
While Expected Threat (xT) \cite{singh_introducing_2018} and VAEP \cite{decroos_actions_2019,decroos_vaep_2021} are prominent in football, the broader principle of valuing actions via expected possession value has been applied across various invasion sports \cite{xarles_action_2025}, highlighting how valuation models must be customized to sport-specific geometries and rules. Methodologically, models diverge between discrete state space representations (e.g., zones or transition matrices in hockey and rugby \cite{routley2015markov,sawczuk2021development}) and continuous court models (e.g., expected possession value in basketball and rugby \cite{cervone2016multiresolution,sawczuk2024bayesian}). Furthermore, models must incorporate dynamic, rule-specific constraints, such as power plays in ice hockey \cite{Liu2018DeepRL}, shot clocks in basketball \cite{sandholtz2020markov}, tackle limits in rugby \cite{sawczuk2021development}, touch phases in volleyball \cite{yoshihara2025play}, and down-and-distance situations in American football \cite{yurko2019nflwar}. 

In handball, player valuation is dominated by traditional statistics or box-score-derived heuristic indices like the HPI \cite{hbl_hpi}, which lack spatial and sequential context. Existing advanced models are either limited to terminal shots, such as Expected Goals (xG) frameworks \cite{adams_expected_2023,mortelier2024what}, or operate on continuous tracking trajectories, such as the spatiotemporal EPV framework PIVOT \cite{muller_pivot_2022}. Consequently, handball lacks a tailored, event-based framework that values discrete actions while incorporating sport-specific geometries and rules (e.g., native court zones and passive play constraints). Furthermore, existing player ratings have not been validated using structured meta-analytics frameworks \cite{davis_methodology_2024,franks_meta-analytics_2016,sportstutorial2024}. We bridge these gaps by adapting and systematically evaluating xT and VAEP for
handball, with H-xT representing the discrete state space models and H-VAEP the
models that additionally condition on the preceding actions and the game
context.

\section{Adapting Expected Threat to Handball}
\label{sec:adapting_xt}
\begin{figure}[t]
    \centering
    \begin{subfigure}[b]{0.32\textwidth}
        \centering
        \includegraphics[width=\textwidth]{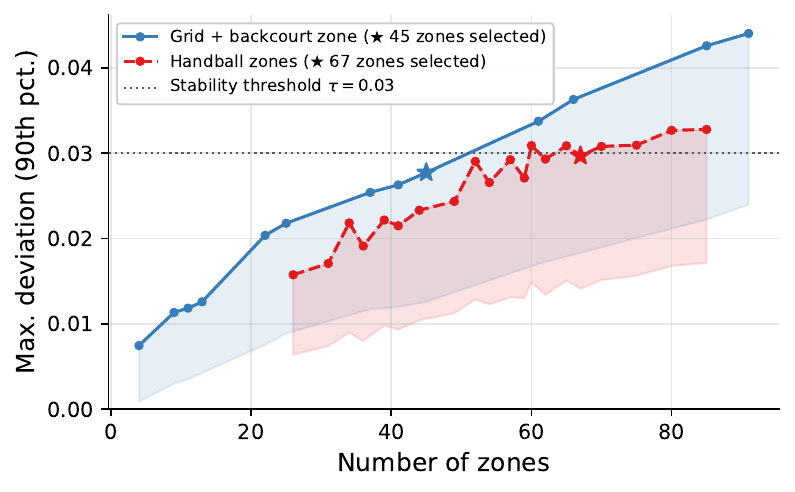}
        \caption{Robustness comparison.}
        \label{fig:xt_stability}
    \end{subfigure}
    \hfill
    \begin{subfigure}[b]{0.32\textwidth}
        \centering
        \includegraphics[width=\textwidth]{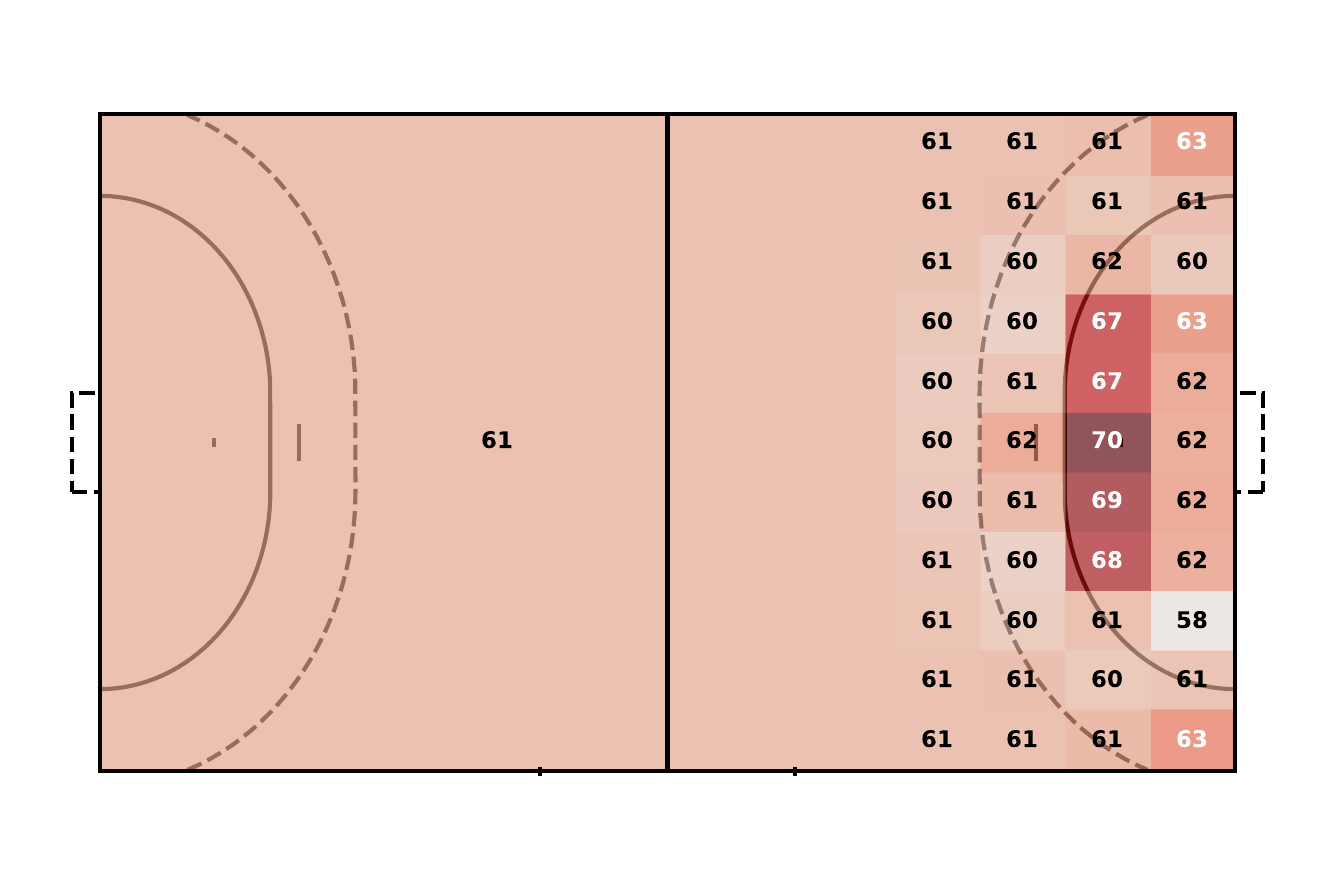}
        \caption{Grid-based xT.}
        \label{fig:xt_grid}
    \end{subfigure}
    \hfill
    \begin{subfigure}[b]{0.32\textwidth}
        \centering
        \includegraphics[width=\textwidth]{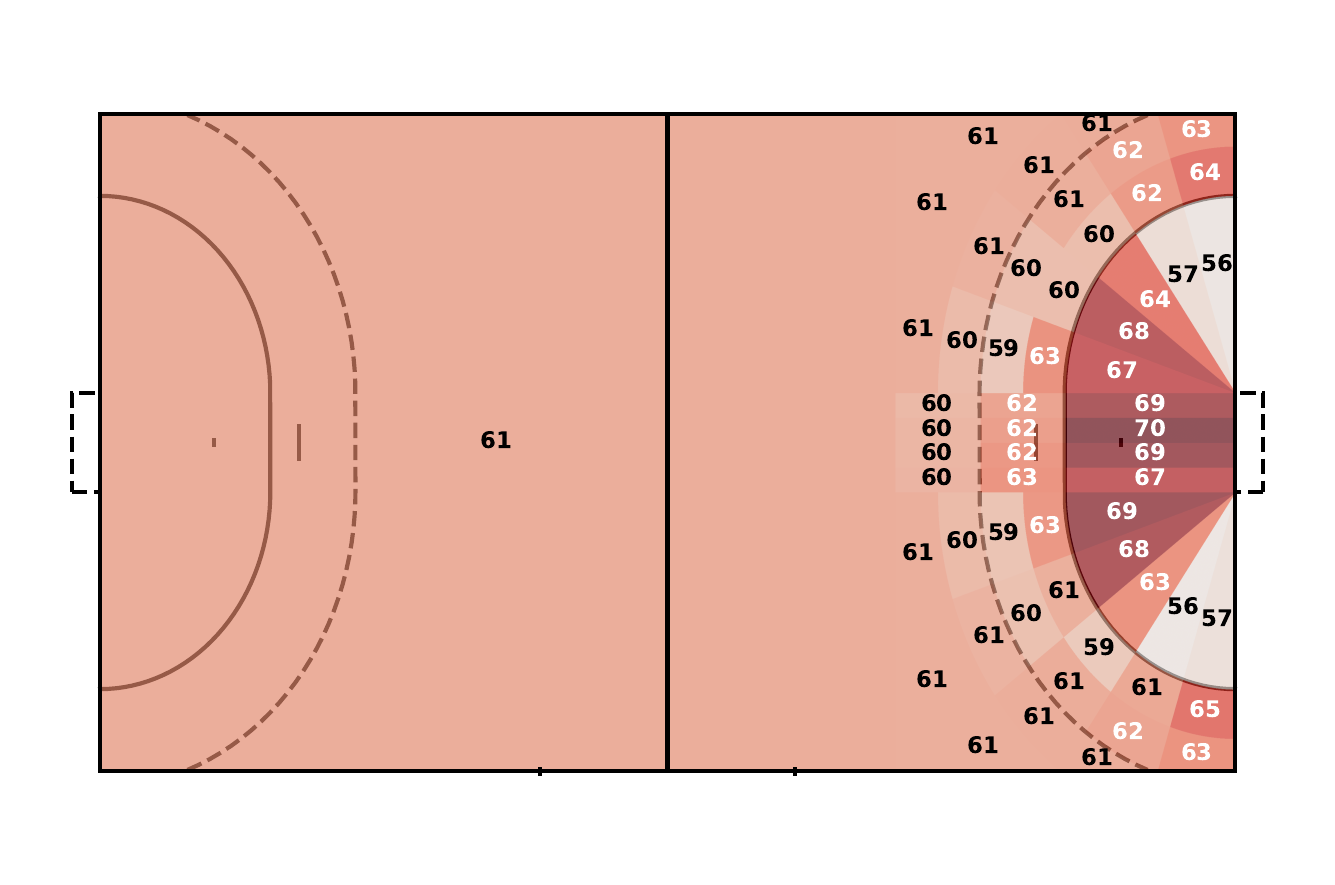}
        \caption{Handball-native xT.}
        \label{fig:xt_handball}
    \end{subfigure}
    \caption{Robustness analysis and Expected Threat surfaces. (a) shows the 90th percentile of maximum deviation (shaded down to the 10th percentile) as a function of zone count for grid-based vs.\ handball-native layouts following~\cite{van_arem_trade-off_2025}. (b) and (c) display the xT surfaces (in \%) fitted on the 2024/25 season.}
    \label{fig:main_2x2_grid}
\end{figure}

The Expected Threat (xT) framework~\cite{rudd_framework_2011,singh_introducing_2018} evaluates ball progression actions using a spatial Markov chain. The court is discretized into zones, and the expected threat value $xT(z)$ represents the probability that a possession currently in zone $z$ leads to a goal. Transitions between zones are determined by movement and shot probabilities, which can be solved recursively as detailed in~\cite{singh_introducing_2018}. Once these zone values are computed, any individual ball progression action starting in zone $z_{\mathrm{start}}$ and ending in $z_{\mathrm{end}}$ is valued by the difference in expected threat: $\Delta xT = xT(z_{\mathrm{end}}) - xT(z_{\mathrm{start}})$. How to discretize the playing area is a critical design choice in xT. While football analytics typically relies on rectangular grids (e.g., a $16 \times 12$ layout), this approach is ill-suited for handball court geometry, which is defined by curved 6m goal creases and 9m free-throw arcs. Consultations with professional coaches confirmed that rectangular cells are unintuitive because grid lines intersect these arcs arbitrarily, destroying tactical interpretability. Consequently, we developed a handball-native court zoning layout (illustrated in Fig.~\ref{fig:xt_handball}) featuring angular boundaries originating from the goals and concentric divisions matching the 6m and 9m lines. Furthermore, since build-up play occurs almost exclusively in the opponent's half, we aggregate the entire defensive half into a single zone. Under a full grid, defensive-half zones converge to virtually identical expected threat values, with minor differences representing spurious noise from action scarcity.

To select the optimal number of zones and compare layouts, we adopt the simulation-based robustness methodology of~\cite{van_arem_trade-off_2025}, which balances the trade-off between model flexibility (capturing fine-grained tactical movements) and robustness (avoiding high parameter variance). We utilize tracking-derived event sequences across five HBL seasons (2021/22 and 2022/23 as development set; 2023/24, 2024/25, and 2025/26 as held-out evaluation set). We apply light data cleaning, including removing the first 30\,s of each match to exclude pre-match ceremonial passing that would skew transition probabilities. We represent the event sequences using a schema adapted from SPADL~\cite{decroos_actions_2019} by omitting body part features and defining four action types: passes, dribbles (ball possession), field shots, and seven-meter penalty shots. Since pass success labels are missing in portions of the tracking data, success is imputed by verifying if the passing team retains possession for the subsequent action. We fit a ground-truth model, $M_{\mathrm{full}}$, using the combined 2021/22 and 2022/23 development seasons. To quantify the robustness-flexibility trade-off, we then simulate $B = 1000$ seasons by sampling 306 matches (representing one HBL season) from the development set. For each simulation $b$, we fit a model $M_b$ and calculate the maximum zone-level absolute deviation from the ground truth: $D_b = \max_z |xT_b(z) - xT_{\mathrm{full}}(z)|$. Following~\cite{van_arem_trade-off_2025}, the robustness metric $R_{90}$ is the 90th percentile of $D_b$ across all simulations. To ensure a fair comparison via the total number of zones as a flexibility measure, we introduce a single backcourt zone in the grid model as well. Fig.~\ref{fig:xt_stability} shows the robustness curves for both layouts. For any number of zones, the grid model's 90th percentile deviation is larger than that of the handball-native model, demonstrating that our native layout is systematically more robust. Applying the decision rule of choosing the most flexible configuration with $R_{90} \le 0.03$~\cite{van_arem_trade-off_2025}, the handball-native layout supports a capacity of 67 zones, whereas the grid model is restricted to 45 zones.

The 2024/25 season xT surfaces (Figs.~\ref{fig:xt_grid} and \ref{fig:xt_handball}) reveal key tactical insights. First, inside the 6m crease, threat increases with the goal angle, reflecting higher expected shot values from central positions. Second, a low-threat trough exists directly outside the 9m line compared to the outer backcourt: as attackers approach the 9m boundary, defensive contact increases turnovers. Once this line is penetrated, threat rises near the central 6m line due to high-quality shooting opportunities. These central zones are likely undervalued as H-xT excludes 7m penalties due to missing foul coordinates. Third, wing corners outside the 6m line exhibit the highest threat outside the crease. Wing passes enable players to jump into the crease where defensive contact is restricted, yielding clean shots with minimal interference.

\section{Adapting VAEP to Handball}
\label{sec:adapting_vaep}
The VAEP framework~\cite{decroos_actions_2019,decroos_vaep_2021} rates on-ball actions by their impact on a team's short-term scoring and conceding probabilities. Specifically, it uses machine learning to estimate $P_{\mathrm{scores}}(S_i)$ and $P_{\mathrm{concedes}}(S_i)$, which represent the probabilities that the team possessing the ball scores or concedes a goal within the next $N = 10$ actions from a given game state $S_i$, represented by the last $k$ actions. An action's value is then defined as the change in scoring probability minus the change in conceding probability caused by it: $V(a_i) = \Delta P_{\mathrm{scores}} - \Delta P_{\mathrm{concedes}}$. We refer the reader to~\cite{decroos_actions_2019} for the full mathematical details of the framework. 

When adapting VAEP to team handball, we optimize the feature space, the learning algorithm, and the context length $k$ for the sport's geometries and high-scoring dynamics. To evaluate different configurations, we train models on the 2021/22 season and evaluate on the 2022/23 season, measuring Brier score for calibration and ROC-AUC for rank ordering. Uncertainty is quantified by a game-clustered bootstrap (1,000 replicates over the 306 test games, drawn once and shared across configurations so that comparisons are paired). We control the family-wise error rate with Holm's correction. Hyperparameters are tuned via 5-fold cross-validation on game-level splits using a tree-structured Parzen estimator (50 trials). The original football-native model serves as our baseline~\cite{decroos_actions_2019} (CatBoost with default parameters, original feature set, $k=3$).

\begin{table}[t]
    \centering
    \caption{Predictive performance (Brier score $\downarrow$, ROC-AUC $\uparrow$) of H-VAEP across configurations of feature sets, algorithms, and context lengths $k$ (trained on 2021/22, evaluated on 2022/23). The `No Features' baseline predicts marginal training set probabilities. Bold entries highlight the baseline configuration (Original VAEP), the best feature combination (1b+2+3), the chosen model (XGBoost, $k=3$), and the best overall model ($k=6$).}
    \label{tab:vaep}
    \scriptsize
    \setlength{\tabcolsep}{4.0pt}
    \begin{tabular}{lccc cc cc}
        \toprule
         & & & & \multicolumn{2}{c}{\textbf{$P_{\mathrm{scores}}$}} & \multicolumn{2}{c}{\textbf{$P_{\mathrm{concedes}}$}}\\
        \textbf{Feature Set} & \textbf{Algorithm} & \textbf{Context} & \textbf{Tuned} & \textbf{Brier} & \textbf{AUC} & \textbf{Brier} & \textbf{AUC} \\
        \midrule
        No Features & - & - & -               & 0.14292 & -       & 0.03008 & - \\
        Original VAEP & CatBoost & 3 & \xmark & \textbf{0.11461} & \textbf{0.78252} & \textbf{0.02780} & \textbf{0.79091} \\
        Original VAEP & CatBoost & 3 & \cmark & 0.11321 & 0.78743 & 0.02751 & 0.79959 \\
        \midrule
        1a) Scores Rem. & CatBoost & 3 & \cmark  & 0.11308 & 0.78855 & 0.02749 & 0.79895 \\
        1b) Scores Repl. & CatBoost & 3 & \cmark & 0.11303 & 0.78894 & 0.02749 & 0.79970 \\
        2) Location Repl. & CatBoost & 3 & \cmark & 0.11316 & 0.78748 & 0.02750 & 0.79991 \\
        3) Context Added & CatBoost & 3 & \cmark & 0.11287 & 0.79121 & 0.02752 & 0.80251 \\
        1b) + 2) + 3) & CatBoost & 3 & \cmark & \textbf{0.11257} & \textbf{0.79338} & \textbf{0.02748} & \textbf{0.80391} \\
        \midrule
        1b) + 2) + 3) & Ran. Forest & 3 & \cmark & 0.11648 & 0.77965 & 0.02811 & 0.78599 \\
        1b) + 2) + 3) & XGBoost & 3 & \cmark & \textbf{0.11220} & \textbf{0.79582} & \textbf{0.02735} & \textbf{0.80683} \\
        \midrule
        1b) + 2) + 3) & CatBoost & 6 & \cmark & 0.11214 & 0.79741 & 0.02742 & 0.80771 \\
        1b) + 2) + 3) & XGBoost & 6 & \cmark & \textbf{0.11174} & \textbf{0.79990} & \textbf{0.02733} & \textbf{0.81007} \\
        \bottomrule
    \end{tabular}
\end{table}

Table~\ref{tab:vaep} shows that applying football-native features directly is suboptimal. We adapt the feature space in three key areas. First, to prevent overfitting from sparse, high-scoring patterns, we replace exact cumulative scores (which generalize poorly in handball's high-scoring environment) with a bucketed score difference (1b: categorized from strongly behind to strongly leading). Second, to represent shooting quality more effectively, we replace the polar angle with the goal angle between the rays connecting the ball to the two posts (2). This visual angle is more critical in handball than in football, as handball attacks allow shots from almost any position during a set play, whereas football possessions feature only brief windows in potential shooting locations. Third, we capture tactical pace by adding the elapsed seconds since gaining possession (3), which distinguishes rapid fast breaks from slower set attacks and indicates upcoming shot pressure under handball's passive play rule. As shown in Table~\ref{tab:vaep}, combining these complementary features (1b+2+3) yields a compounding performance improvement. All individual and the combined feature modifications improve both Brier score and ROC-AUC significantly ($\alpha=0.05$, Holm corrected) over the tuned baseline for $P_{\mathrm{scores}}$, with the exception of the location replacement, whose ROC-AUC gain is not significant. For $P_{\mathrm{concedes}}$ the effects are weaker and split by metric: no feature modification significantly improves the Brier score, while the added context feature (3) and the combined set do improve ROC-AUC significantly (both Holm corrected $p<0.05$). The adaptations therefore sharpen the ranking of conceding risk without measurably improving its calibration.

In terms of model selection, XGBoost achieves the best performance. Notably, the gain from customizing the features ($+0.00596$ scoring ROC-AUC over the tuned CatBoost baseline, 95\% CI $[0.00509, 0.00678]$) is more than double the gain from subsequent model class selection ($+0.00244$ scoring ROC-AUC, $[0.00211, 0.00276]$).

\begin{figure}[t]
    \centering
    \begin{subfigure}[b]{0.47\textwidth}
        \centering
        \includegraphics[width=\textwidth]{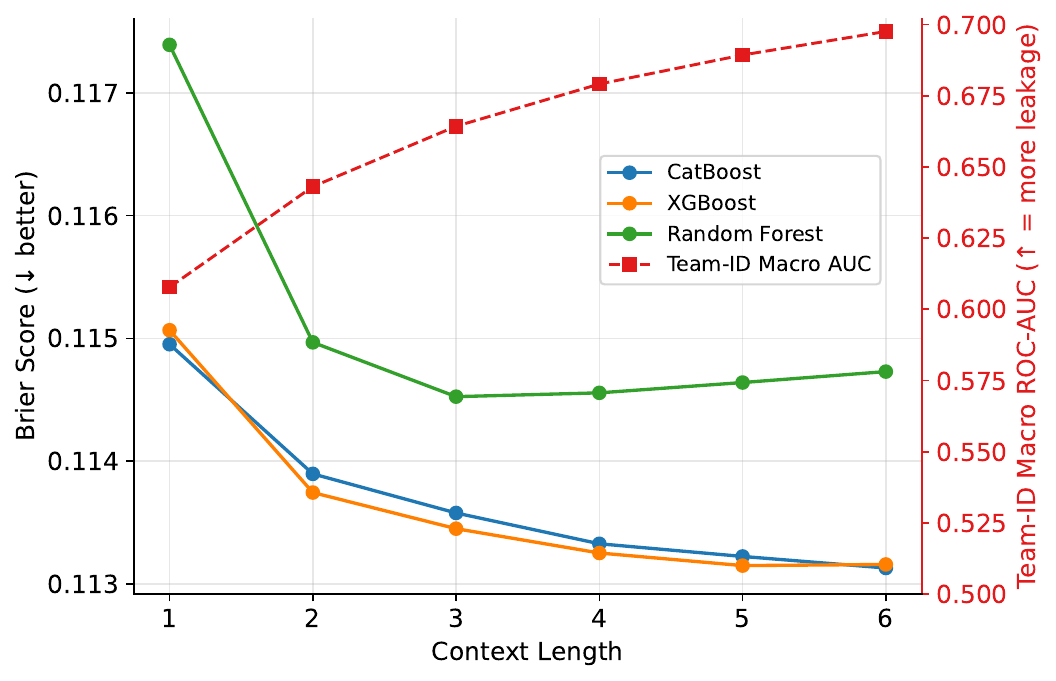}
        \caption{Scoring target ($P_{\mathrm{scores}}$).}
        \label{fig:brier_scores}
    \end{subfigure}%
    \hfill
    \begin{subfigure}[b]{0.47\textwidth}
        \centering
        \includegraphics[width=\textwidth]{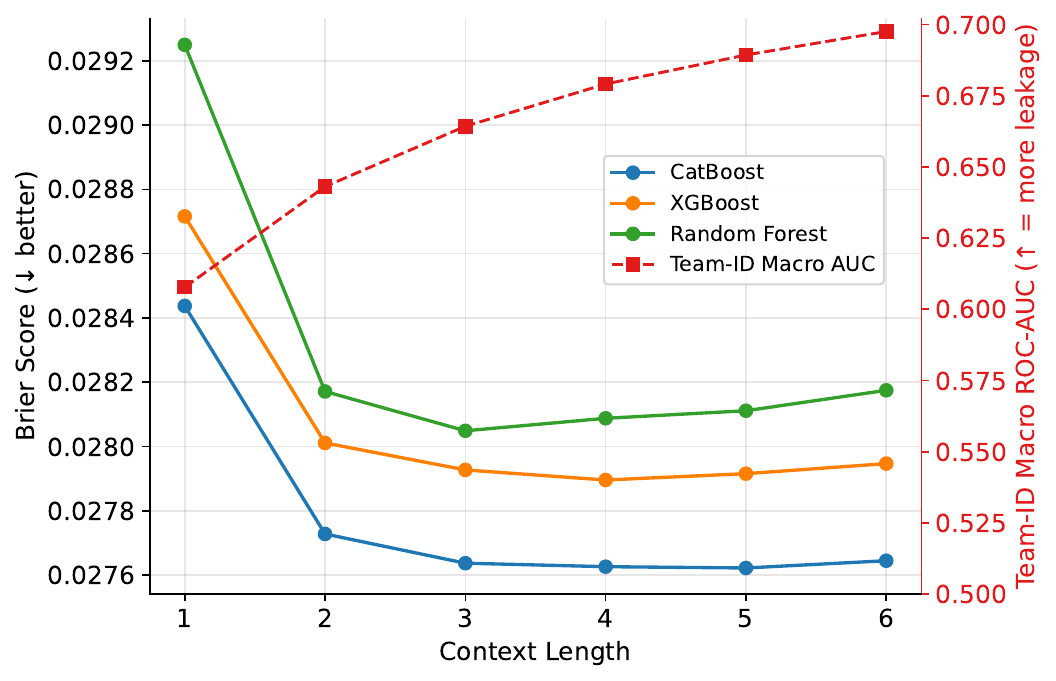}
        \caption{Conceding target ($P_{\mathrm{concedes}}$).}
        \label{fig:brier_concedes}
    \end{subfigure}
    \caption{Predictive performance (Brier score, left axis, solid) and team-identity leakage (Macro-AUC, right axis, dashed) across context lengths $k$.}
    \label{fig:brier_and_leakage}
\end{figure}

While extending possession history (context length $k$) provides more sequential context to capture tactical patterns, it increases the risk of overfitting and team-identity leakage, which occurs when a model identifies team-specific playing styles and predicts outcomes based on team identity rather than general action quality~\cite{sportstutorial2024}. To quantify this, we train a classifier to predict the identity of the team executing the action from its sequence context, measuring its Macro-AUC on a stratified 20\% game-level holdout within the 2021/22 season. As shown in Fig.~\ref{fig:brier_and_leakage}, team leakage increases monotonically with context length, rising from 0.608 at $k=1$ to 0.664 at $k=3$ and 0.698 at $k=6$. Although extending context to $k=6$ improves predictive performance (Table~\ref{tab:vaep}), it incurs substantially higher team-identity leakage. Thus, we select $k=3$ as a balanced compromise between predictive quality and team-identity bias. Compared to the original VAEP baseline, our final model configuration (tuned XGBoost, 1b+2+3, $k=3$) reduces the scoring Brier score to 0.11220 ($-0.00241$, 95\% CI
$[-0.00287, -0.00205]$) and the conceding Brier score to 0.02735 ($-0.00045$, $[-0.00054, -0.00037]$), while improving scoring ROC-AUC to 0.79582 ($+0.01330$, $[0.01174, 0.01491]$) and conceding ROC-AUC to 0.80683 ($+0.01592$, $[0.01311, 0.01876]$).

\begin{figure}[tb]
    \centering
    \begin{subfigure}[b]{0.47\textwidth}
        \centering
        \includegraphics[width=\textwidth]{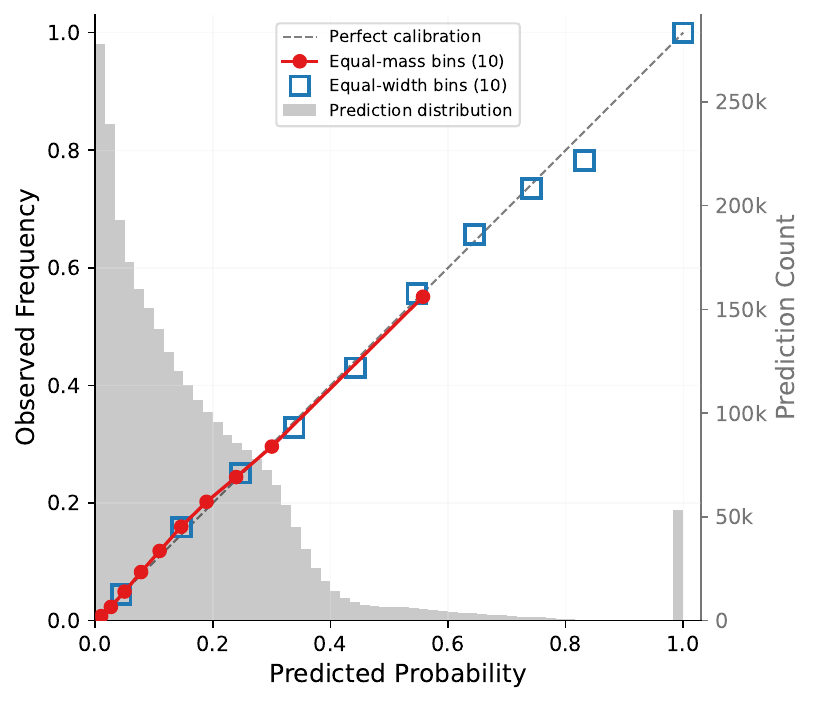}
        \caption{Scoring target ($P_{\mathrm{scores}}$).}
        \label{fig:calibration_scores}
    \end{subfigure}%
    \hfill
    \begin{subfigure}[b]{0.47\textwidth}
        \centering
        \includegraphics[width=\textwidth]{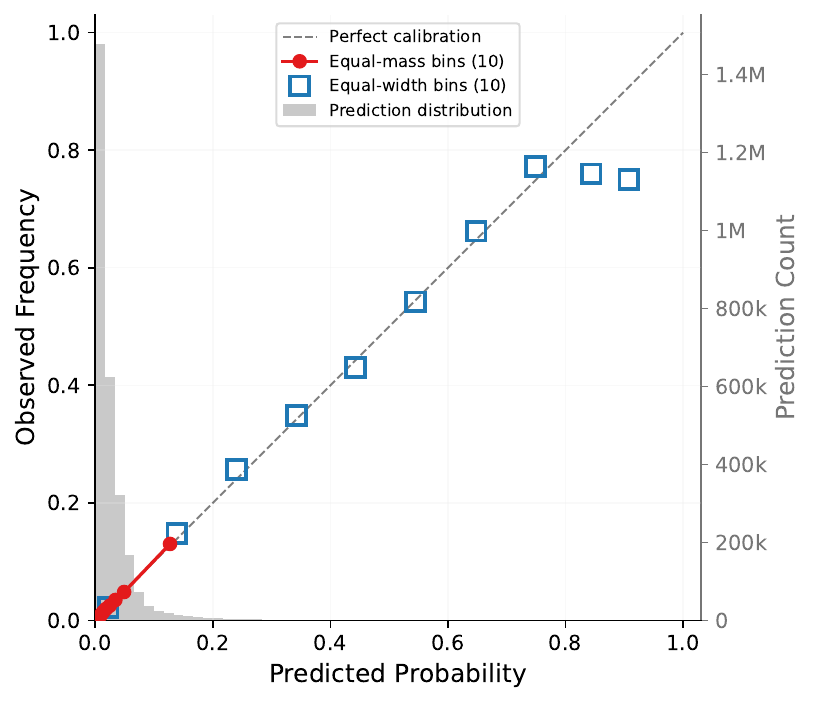}
        \caption{Conceding target ($P_{\mathrm{concedes}}$).}
        \label{fig:calibration_concedes}
    \end{subfigure}
    \caption{Reliability of the final H-VAEP model, pooled over the three held-out seasons 2023/24, 2024/25, and 2025/26. Red dots mark equal-mass decile bins, blue squares equal-width bins, and the gray histogram the distribution of the predicted probabilities (right axis).}
    \label{fig:calibration}
\end{figure}

We also assess the final model configuration's performance and calibration on the three held-out seasons (2023/24 to 2025/26) with a rolling training and prediction setup: models are trained on season $n-1$ (starting with 2022/23) to compute action valuations for season $n$ so that every reported probability is out of sample. Pooled over the 2.83M valued game states, the scoring model attains a Brier score of 0.11341 at a ROC-AUC of 0.79128 and the conceding model a Brier score of 0.02654 at a ROC-AUC of 0.79454, against base rates of 17.3\% and 3.0\%. In addition, Fig.~\ref{fig:calibration} reports the
reliability of both classifiers. The scoring model follows the diagonal closely over the entire populated range (expected calibration error 0.006 over decile bins), including the rare predictions above 50\%, indicating that it is well calibrated. Its largest systematic deviation is an underestimation of 1.4 percentage points around 15\%, and the apparent overestimation in the 80--90\% bin rests on just 1,733 of 2.83M predictions. The spike of the histogram at 100\%, and the exactly calibrated equal-width bin above 90\%, are the 53k successful shots (1.9\% of the states), whose scoring probability we replace deterministically by 100\% because a goal fixes the outcome of the state by definition. The conceding model is calibrated equally well (expected calibration error 0.001) and stays on the diagonal up to 75\% despite how seldom such states occur. Only above 80\%, where fewer than 0.02\% of the predictions lie, does it overestimate the risk of conceding.

\section{Empirical Player Ratings and Value Composition}
\label{sec:resulting_ratings}

To evaluate player ratings for the 2024/25 season, we restrict the analysis to players with at least 500 total minutes and 250 minutes in offense, using official, human-verified box-score statistics from Sportradar and HPI data from the HBL website for baseline comparison. The 500-minute threshold represents a comparable standard to the 900 minutes used in football VAEP~\cite{decroos_actions_2019}, reflecting handball's higher physical intensity and rotation rates where players rarely exceed 50 minutes per match. The additional offensive playing time filter is necessary to exclude defensive specialists, as the tracking system's automatically recognized event data do not yet include defensive events. 

The resulting rankings demonstrate strong face validity: the ten players with the highest H-VAEP per 10 minutes in offense (H-VAEP/10o) are widely recognized elite performers who, between 2022 and 2025, collectively received three IHF World Handball Player of the Year awards, three HBL MVP titles, two EHF Champions League Final Four MVP awards, and two HBL Best Young Player awards. At the same time, the ranking is not a mere reproduction of box-score statistics. The player ranked first in H-VAEP/10o ranks only 55th in goals and 62nd in goals/10o, 9th in assists, 8th in assists/10o, and 27th in HPI: traditional statistics do not fully capture this player's exceptional build-up play, which our model highlights by valuing the complete action sequence. Conversely, the player leading the league in both goals/10o and assists/10o ranks sixth in H-VAEP/10o. While goals and assists capture the terminal actions of possessions, H-VAEP incorporates the value of the preceding build-up play and accounts for actions like turnovers, providing a more comprehensive view of offensive contributions beyond box-score totals.

\begin{figure}[tb]
    \centering
    \includegraphics[width=\textwidth]{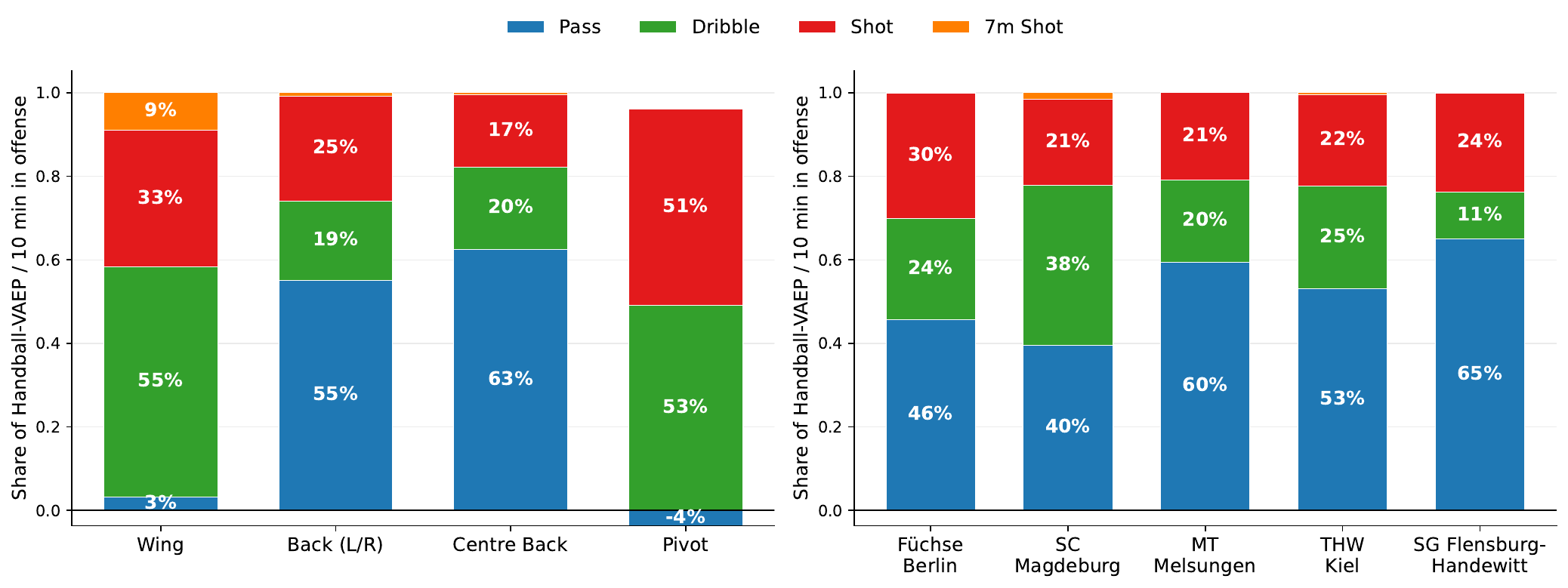}
    \caption{Relative Handball-VAEP decomposition by action type in the 2024/25 season. The left panel shows the composition of the average H-VAEP per 10 minutes in offense for each position group. The right panel shows it for the back players (left, right, and centre backs) of the season's top five teams. Averages are weighted by offensive playing time.}
    \label{fig:vaep_decomposition}
\end{figure}

Fig.~\ref{fig:vaep_decomposition} presents the action-type decomposition of H-VAEP/10o aggregated by position group and team. The left panel reflects the distinct offensive roles of the positions. Back players generate the majority of their value through passing (55\% for left and right backs, 63\% for centre backs), consistent with their playmaking responsibilities. Wings instead accumulate most of their value through dribbles (55\%) and shots (33\%), as they finish fast breaks after carrying the ball over long distances and attack the crease from the narrow angle of their corner position. Their passing contributes almost nothing (3\%), since passes from the wing typically return the ball to the backcourt and rarely improve the attacking situation. Pivots derive their value from shots (51\%) and dribbles (53\%), while their passing contribution is slightly negative ($-4$\%). A pivot who receives the ball at the six-meter line and cannot turn toward the goal usually plays the ball backward, which keeps the attack going but hardly improves the scoring opportunity, and playing out of this congested space additionally carries a considerable turnover risk.

The right panel shows that the value composition within the same position group differs markedly between teams. SC Magdeburg's tactical system heavily emphasizes isolation plays and breakthrough runs, and this is clearly reflected in the ratings. Their back players generate 38\% of their value through dribbles, by far the largest share among the top five teams. SG Flensburg-Handewitt's backs show the opposite profile with the highest share of value through passing (65\%), which results from their emphasis on fast breaks and a system that relies less on isolation plays and breakthroughs.

\begin{figure}[tb]
    \centering
    \begin{subfigure}[b]{0.47\textwidth}
        \centering
        \includegraphics[width=\textwidth]{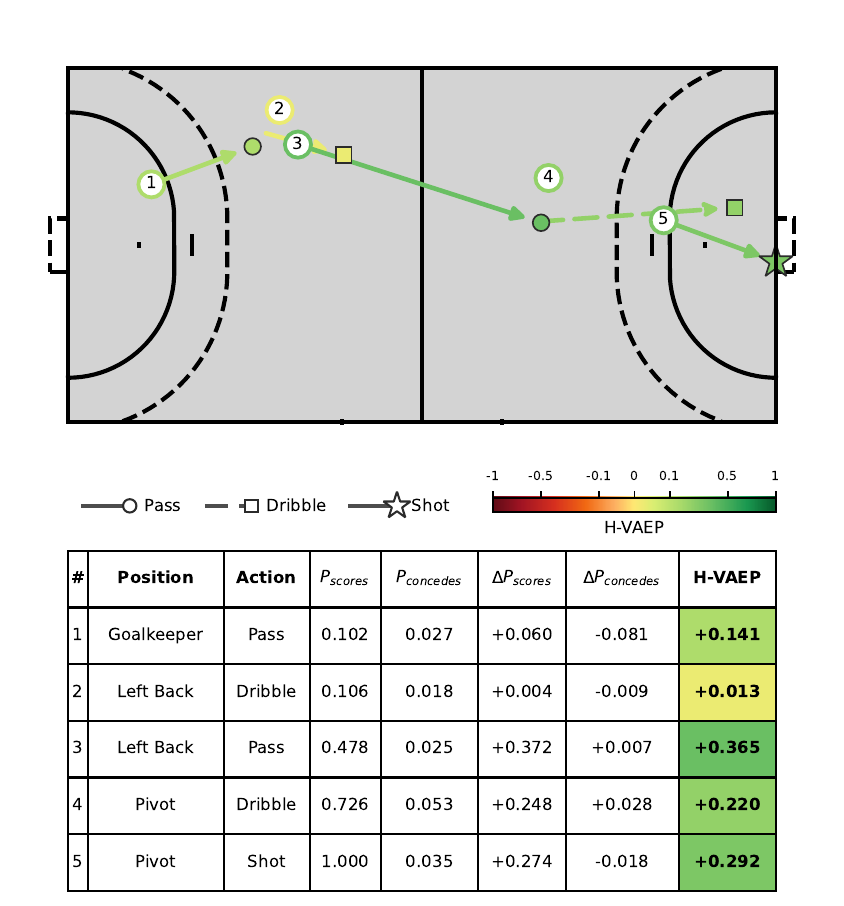}
        \caption{Fast break.}
        \label{fig:action_sequence_fast_break}
    \end{subfigure}%
    \hfill
    \begin{subfigure}[b]{0.47\textwidth}
        \centering
        \includegraphics[width=\textwidth]{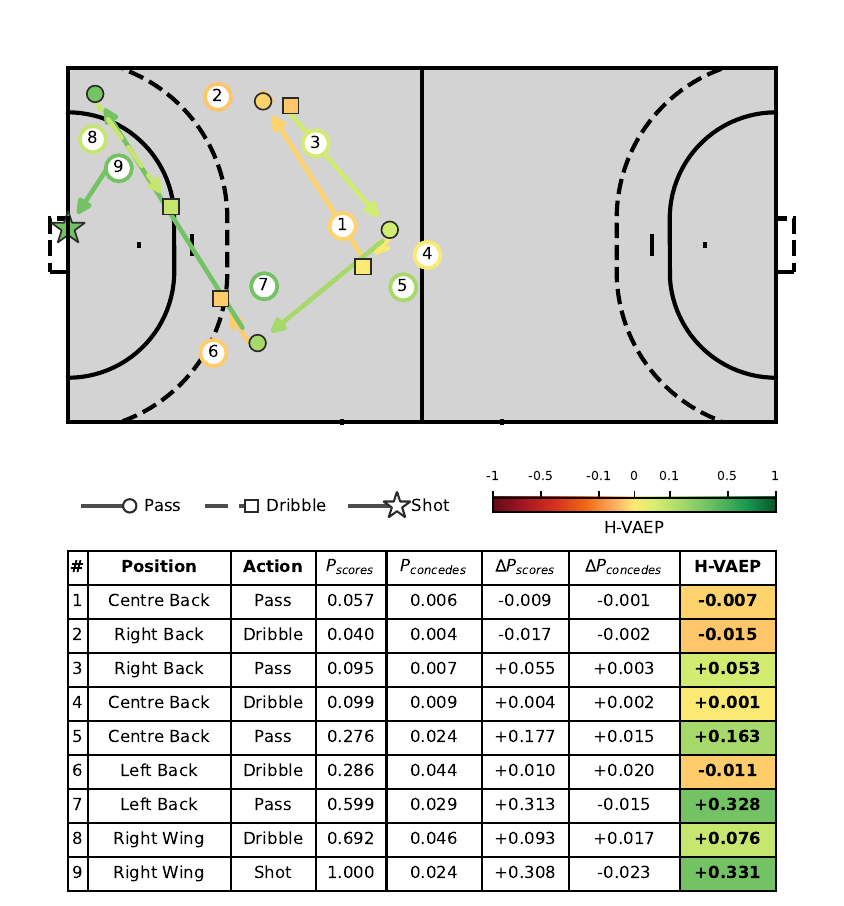}
        \caption{Positional attack.}
        \label{fig:action_sequence_build_up}
    \end{subfigure}
    \caption{H-VAEP valuation of two action sequences by SG Flensburg-Handewitt in the 2024/25 season. Arrow color encodes an action's H-VAEP, line style and end marker its type, and the numbers its order in the sequence.}
    \label{fig:action_sequence}
\end{figure}

Finally, we illustrate the face validity of the valuations themselves on two sequences by SG Flensburg-Handewitt from the 2024/25 season
(Fig.~\ref{fig:action_sequence}). In the fast break (Fig.~\ref{fig:action_sequence_fast_break}), H-VAEP rates the left back's long pass as the most valuable action ($+0.365$) and splits the pivot's contribution between the dribble that improves the shooting position ($+0.220$) and the shot itself ($+0.292$). The pivot therefore keeps the credit for creating the opportunity even if the shot is missed, in which case only the shot is penalized. The goalkeeper receives $+0.141$ for the outlet pass that starts the break before the defense is set, a contribution that box-score statistics and
the HPI ignore entirely. The positional attack (Fig.~\ref{fig:action_sequence_build_up}) shows that the model separates passes that look alike. The centre back's slow ball to the right back is rated neutral ($-0.007$), the faster return pass slightly positive ($+0.053$), and the hard
pass into the run of the left back clearly positive ($+0.163$) as it initiates the attack. The left back
then turns a contested 9m shooting position into an uncontested wing shot with a diagonal ball ($+0.328$). In total, the centre back and left back receive more credit for creating the chance ($+0.481$) than the wing for finishing it ($+0.407$), reflecting how H-VAEP distributes value across a build-up sequence.

\section{Evaluation of Action Valuation Metrics}
\label{sec:rating_evaluation}
\begin{figure}[b]
    \centering
    \begin{subfigure}[b]{0.32\textwidth}
        \centering
        \includegraphics[width=\textwidth, height=3.5cm, keepaspectratio]{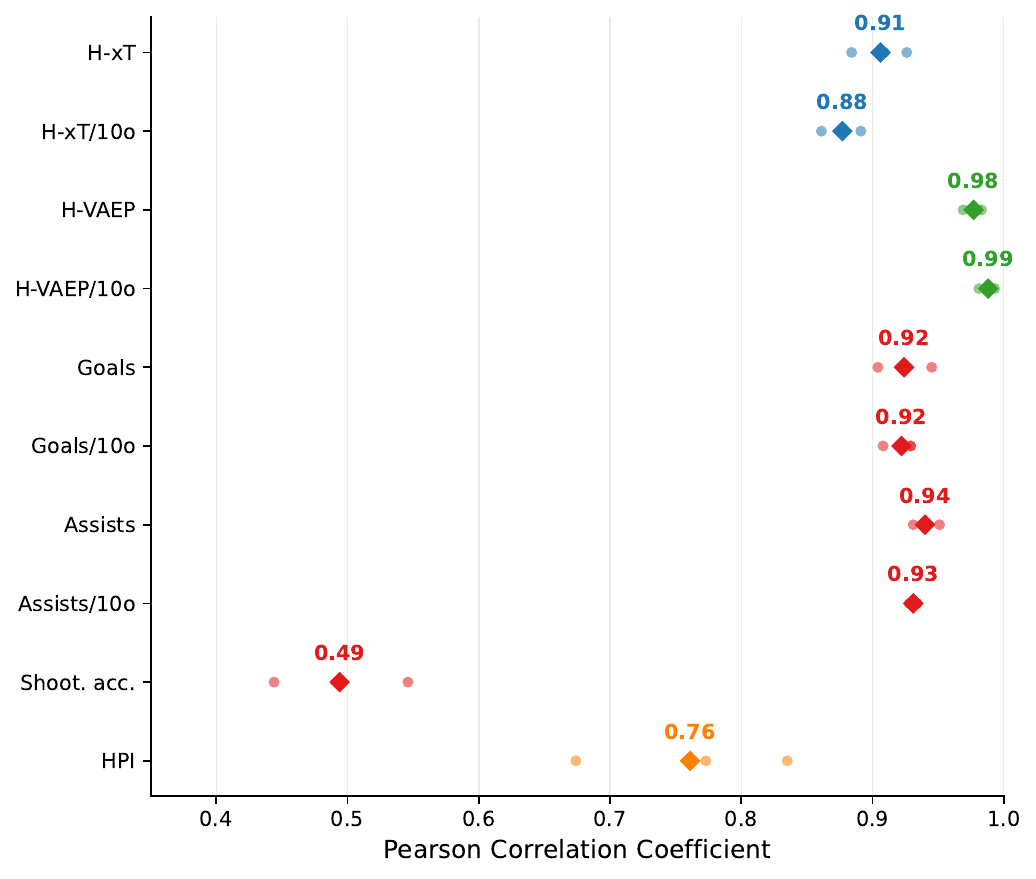}
        \caption{Within-season.}
        \label{fig:within-season_correlations}
    \end{subfigure}
    \hfill
    \begin{subfigure}[b]{0.32\textwidth}
        \centering
        \includegraphics[width=\textwidth, height=3.5cm, keepaspectratio]{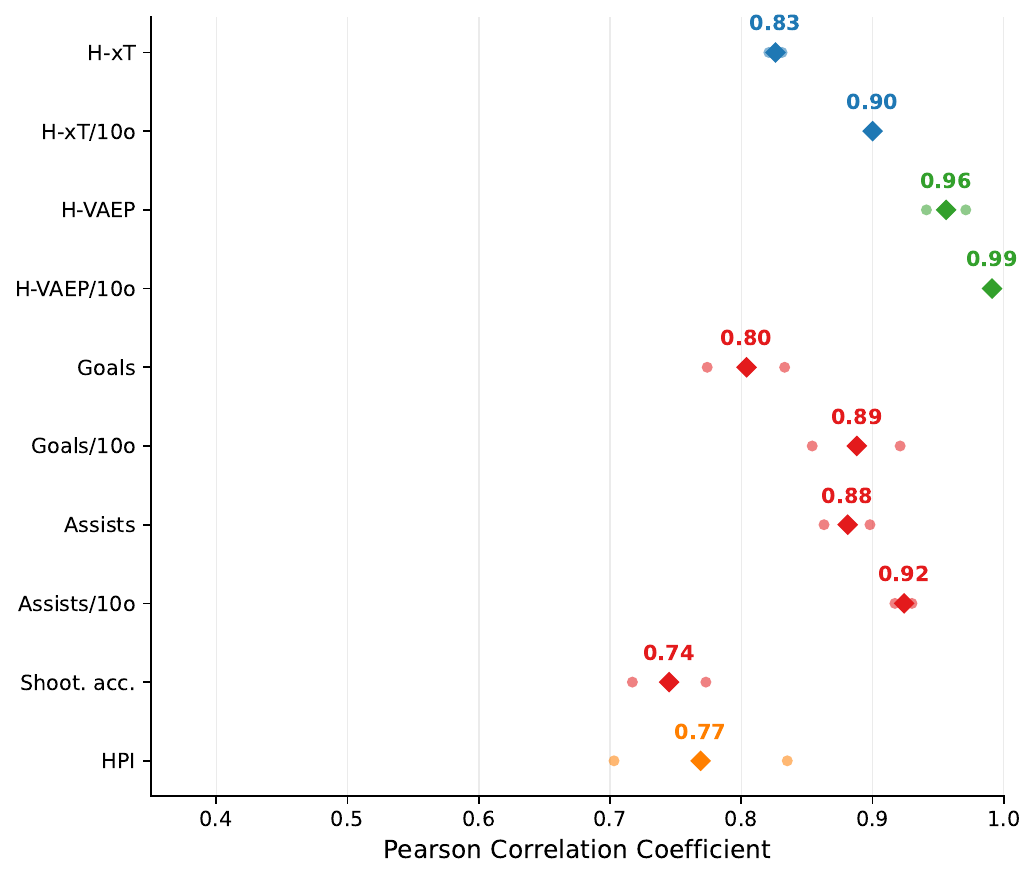}
        \caption{Cross-season.}
        \label{fig:cross-season_correlations}
    \end{subfigure}
    \hfill
    \begin{subfigure}[b]{0.32\textwidth}
        \centering
        \includegraphics[width=\textwidth, height=3.5cm, keepaspectratio]{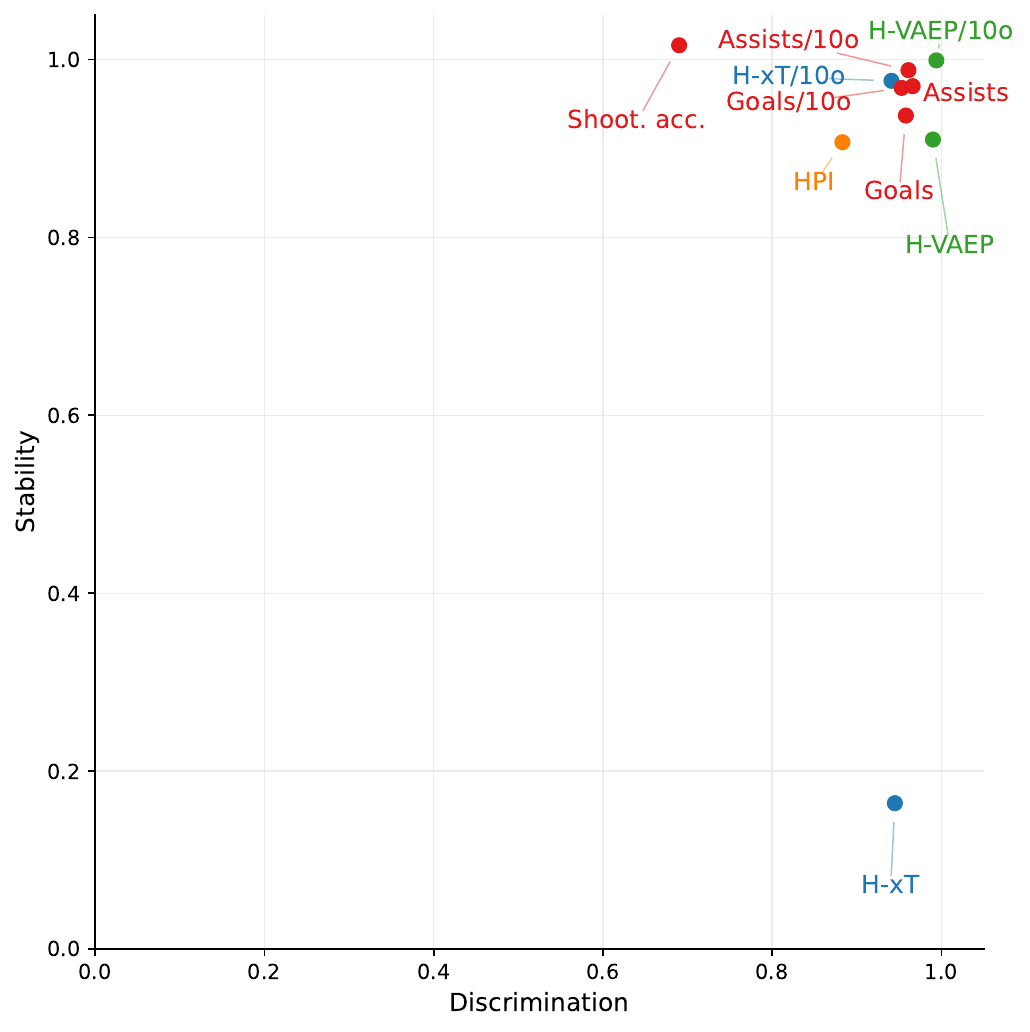}
        \caption{Meta-metrics.}
        \label{fig:meta-metrics}
    \end{subfigure}
    \caption{Reliability and meta-metrics for different player ratings across 2023/24, 2024/25, and 2025/26. Diamonds in (a) and (b) indicate the mean correlation. (c) displays the stability and discrimination meta-metrics from~\cite{franks_meta-analytics_2016}.}
    \label{fig:stability_correlations}
\end{figure}

In addition to the demonstrated face validity, we evaluate the quality, reliability, and utility of H-xT and H-VAEP using a comprehensive validation framework following established sports analytics methodologies~\cite{davis_methodology_2024,franks_meta-analytics_2016,sportstutorial2024}. We rate the three held-out seasons 2023/24 to 2025/26 under the player selection criteria of Section~\ref{sec:resulting_ratings}. Each season's actions are valued
by a model fitted only on the preceding season, which mirrors how a club would deploy the models and ensures that no model was trained on the season it values.

Following~\cite{davis_methodology_2024}, we measure within-season reliability via the Pearson correlation ($r$) between two sets of randomly distributed game days (for players with $\ge 3$ games in each set) and cross-season reliability using consecutive season pairs. As shown in Figs.~\ref{fig:within-season_correlations} and \ref{fig:cross-season_correlations}, traditional counting statistics like goals and assists are highly stable within-season ($r \ge 0.92$, cross-season $r \ge 0.80$), whereas shooting accuracy is the most unstable ($r = 0.49$ within-season, though $r = 0.74$ cross-season). Our proposed H-VAEP and H-VAEP/10o ratings exhibit exceptional reliability (within-season $r \ge 0.98$, cross-season $r \ge 0.96$). This high stability is likely driven by handball's high-scoring nature, which yields a high volume of actions, especially assists and goals, and reduces statistical noise compared to football. 

To evaluate player separation, we compute the discrimination and stability meta-metrics from Franks et al.~\cite{franks_meta-analytics_2016} (Fig.~\ref{fig:meta-metrics}). H-VAEP/10o performs best, leading in both discrimination ($0.994$) and stability ($0.999$). Without time normalization, absolute H-VAEP retains high discrimination ($0.990$) but lower stability ($0.910$). Notably, while absolute H-xT has low single-game stability ($0.164$), normalizing for playing time (H-xT/10o) raises it to $0.976$. This confirms that adjusting for offensive minutes yields more stable player ratings across all metrics.

To determine if our metrics capture novel insights, we analyze their correlation with traditional indicators and compute the independence meta-metric from~\cite{franks_meta-analytics_2016}. H-xT correlates strongly with assists ($r=0.78$, and $r=0.73$ for H-xT/10o with assists/10o), indicating that progression into high-threat zones (e.g., wing crease jumps) often leads to assists. H-VAEP and H-VAEP/10o correlate strongly with goals ($r=0.72$, $r=0.66$) and goals/10o ($r=0.75$, $r=0.77$), respectively. Consequently, both H-VAEP ($0.147$) and H-VAEP/10o ($0.163$) score low on the independence metric (where lower values denote higher correlation with other indicators), similar to goals ($0.178$). Conversely, shooting accuracy ($0.625$) and HPI ($0.377$) exhibit stronger independence. For HPI, this is a direct design choice: it aggregates defensive and disciplinary actions (e.g., blocks, steals, suspensions) that are excluded from our offensive, on-ball schema.

Qualitatively, coaches confirmed that our handball-specific zoning layout is markedly more intuitive than rectangular grids, aligning with their tactical terminology. Furthermore, discussions during a workshop with coaches from the first and second divisions of the Handball Bundesliga showed that they intuitively grasped how the H-VAEP framework decomposes and attributes value across multi-player build-up sequences, validating its practical utility for player assessment.

\section{Conclusion and Future Work}
\label{sec:conclusion}
In this work, we adapted, optimized, and evaluated the Expected Threat (H-xT) and VAEP (H-VAEP) frameworks for professional team handball.
By addressing sport-specific requirements, we developed a handball-native court zoning layout that respects the sport's unique geometry, which was shown to be systematically more robust than standard rectangular grids.
Furthermore, we customized the feature space of the VAEP framework to accommodate handball's high-scoring dynamics and rapid pace, and we selected an optimal context length to balance predictive quality against team-identity leakage.
Our empirical evaluation across three held-out seasons of the Handball Bundesliga demonstrated that our models generate stable, reliable, and intuitive player ratings that align with expert assessments, while successfully crediting non-terminal actions and build-up play.

Several avenues remain for future research.
First, while professional coaches highlight the importance of defensive contributions, defensive actions are not yet automatically recognized by tracking systems, requiring methods to extract defensive events from raw tracking data.
Second, handball performance valuation and credit attribution should expand to off-ball behavior. Valuing spatial denial in defense or gap creation in offense requires processing player trajectories rather than discrete event data.
Third, although possession-based prediction targets align closely with handball's tactical structure, implementing them requires addressing event-detection noise and false positives that currently prevent the robust identification of possession boundaries.


\bibliographystyle{splncs04}
\bibliography{bibliography}

\end{document}